\documentclass[letterpaper, 10 pt, conference]{ieeeconf}  

\IEEEoverridecommandlockouts                              

\usepackage{graphics} 
\usepackage{graphicx}
\usepackage{multirow}
\usepackage{amsmath}
\usepackage{tabu}
\let\labelindent\relax
\usepackage{enumitem,kantlipsum}
\usepackage{pifont}
\usepackage{amssymb}
\usepackage{eucal}
\usepackage{tabularx}

\title{\LARGE \bf
SDVTracker: Real-Time Multi-Sensor Association and Tracking for Self-Driving Vehicles
}

\author{Shivam Gautam, Gregory P. Meyer, Carlos Vallespi-Gonzalez and Brian C. Becker\\
Uber Advanced Technologies Group\\
{\tt\small \{sgautam,gmeyer,cvallespi,bbecker\}@uber.com}}
\begin{document}

\maketitle
\thispagestyle{empty}
\pagestyle{empty}

\begin{abstract}
Accurate motion state estimation of Vulnerable Road Users (VRUs), is a critical requirement for autonomous vehicles that navigate in urban environments. Due to their computational efficiency, many traditional autonomy systems perform multi-object tracking using Kalman Filters which frequently rely on hand-engineered association. However, such methods fail to generalize to crowded scenes and multi-sensor modalities, often resulting in poor state estimates which cascade to inaccurate predictions. We present a practical and lightweight tracking system, SDVTracker, that uses a deep learned model for association and state estimation in conjunction with an  Interacting Multiple Model (IMM) filter. The proposed tracking method is fast, robust and generalizes across multiple sensor modalities and different VRU classes. In this paper, we detail a model that jointly optimizes both association and state estimation with a novel loss, an algorithm for determining ground-truth supervision, and a training procedure. We show this system significantly outperforms hand-engineered methods on a real-world urban driving dataset while running in less than 2.5 ms on CPU for a scene with 100 actors, making it suitable for self-driving applications where low latency and high accuracy is critical. 

\end{abstract}

\section{Introduction}

Self-Driving Vehicles (SDVs) depend on a robust autonomy system to perceive actors and anticipate future actions in order to accurately navigate the world. Interacting well with Vulnerable Road Users (VRUs) \cite{world2009global} such as pedestrians and bicyclists requires good motion estimates. A classical autonomy system that uses structured prediction for actor trajectory prediction \cite{djuric2020wacv, hong2019rules, cui2019icra} needs not only high detection rates to identify objects in the scene, but also robust tracking performance to estimate the motion state. Probabilistic tracking using filters can be a reliable method to estimate the motion state \cite{thrun2002probabilistic}. These methods attempt to refine the motion estimates of previously tracked objects by associating them with a given set of detections in the scene at the current timestamp. 

Failures in association cause inaccurate state estimates, often leading to cascading errors in future associations, state estimations, and trajectory predictions resulting in improper autonomy behavior \cite{ross2011reduction}. In simple scenes, engineered solutions do well. However, associating VRUs in crowded, urban environments is challenging due to occlusions, crowd density, variying motions and intermittent detector false positives or false negatives. Any errors in association break the strict assumption for probabilistic filtering regarding observations belonging to the same actor, leading to egregious errors. Incorporating detectors for additional sensor modalities, such as LiDAR and camera detectors, improves overall recall, but increases the likelihood of mis-association, especially as each sensor has different failure modes and noise characteristics. Learned approaches offer improved performance, but are often restricted to the 2D image plane \cite{sadeghian2017tracking}, require a fixed number of objects \cite{farazi2017online}, need expensive feature extraction on specialized GPU hardware \cite{zhang2019robust}, or can run only offline \cite{zhang2008global}, making them unsuitable to self-driving applications.

\begin{figure}[t!]
  \includegraphics[width=0.485\textwidth]{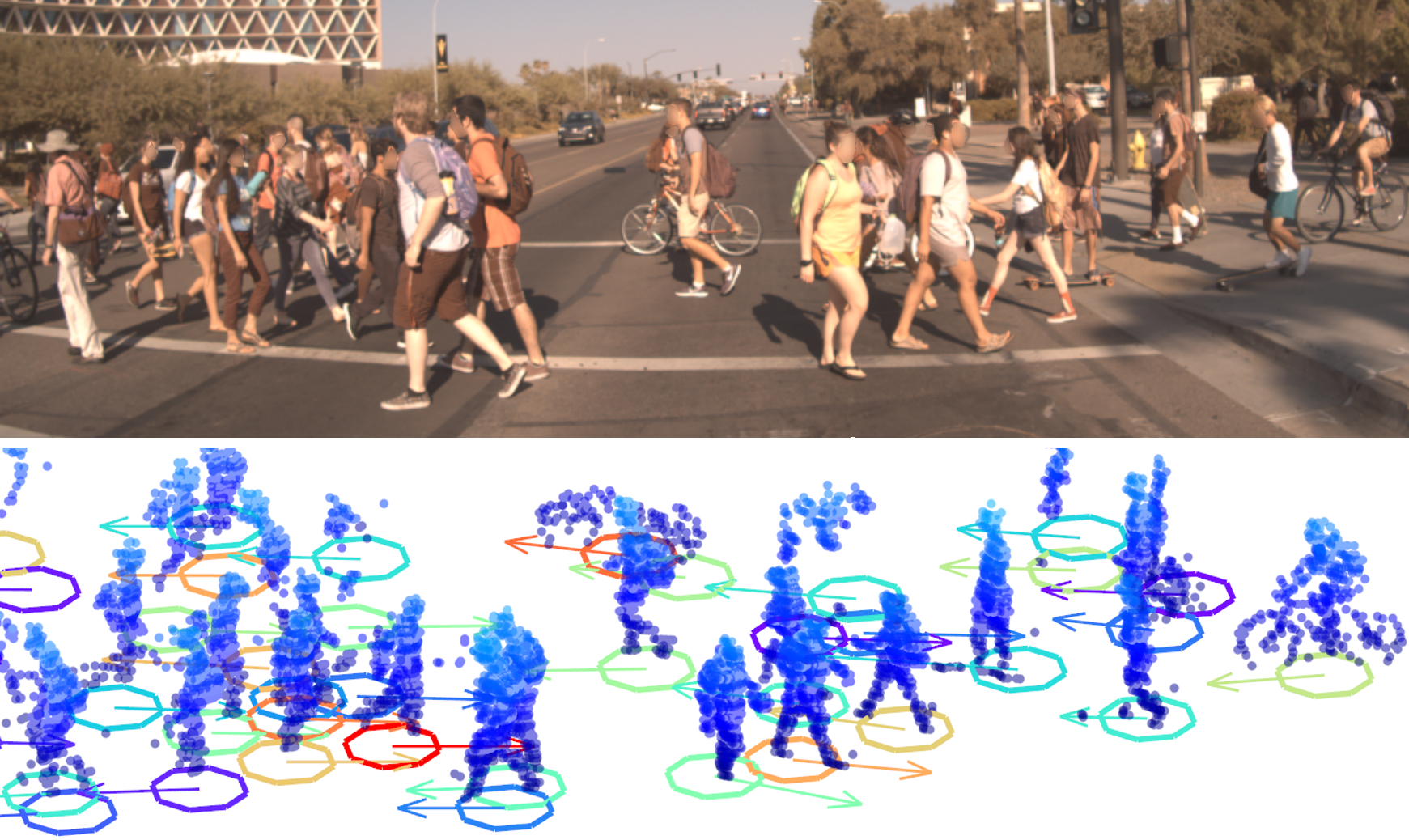}
  \caption{Association and tracking of pedestrians is challenging in dense, urban environments. We propose a real-time learned association and tracking system with IMM filtering that incorporates LiDAR + camera modalities that improves both on the task of association and state estimation.}
  \label{fig:problem_motivation}
\end{figure}

To address these limitations, we propose SDVTracker, a learned association and tracking system for improving motion estimation of VRUs in challenging, self-driving domains. Fig. \ref{fig:problem_motivation} demonstrates our approach performing well in dense crowds across many classes of VRUs including pedestrians, bicyclists, and skateboarders. As the number of VRUs in the scene increases, we show that this method scales better than classical approaches. Our approach generalizes to multi-sensor tracking, improving recall and tracking when both LiDAR and camera detections are used as asynchronous input. In addition to learning association, we propose a novel method to jointly estimate association and state, which leads to improved performance. Further, we show a method of incorporating our learned association and state within a tracking system that uses an Interacting Multiple Model (IMM) filter. Finally, SDVTracker offers real-time performance on commodity CPUs, making it well-suited for compute-limited platforms. 

\begin{figure*}[ht]
  \includegraphics[width=1.0\textwidth]{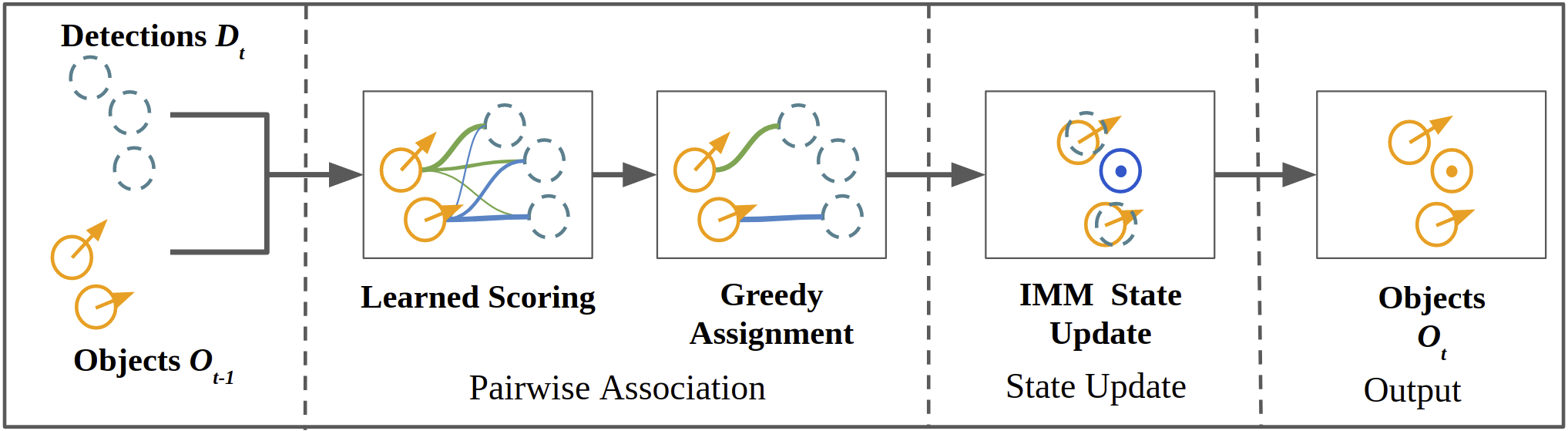}
  \caption{Overview of the association and tracking system. SDVTracker scores the candidate pairs with a learned model to estimate the association probability. After enforcing 1-to-1 correspondence through greedy assignment, we use the learned associations and motion estimates as observations within an IMM update.}
  \label{fig:system_overview}
\end{figure*}

\section{Related Work}
As more autonomous capabilities are added to vehicles, it is critical for these intelligent vehicles to understand and predict the behavior of humans that they interact with to operate safely.
Ohn-Bar and Trivedi \cite{ohn2016looking} provide a thorough survey into three areas of active research where humans and automated vehicles interact, including humans inside the intelligent vehicle, humans around the vehicle, and humans operating surrounding vehicles.
In this work, we focus on understanding the motion of humans around the SDV.

\subsection{Filter-based Tracking} \label{filtering_methods}
A conventional algorithm to perform the motion state estimation from observations is the Kalman Filter (KF) \cite{maybeck1982stochastic}. This algorithm works in two steps that get applied recursively: \emph{prediction} and \emph{update}. In the \emph{prediction} step, the filter produces estimates of the state variables and their uncertainties. The \emph{update} step is performed when the new measurement arrives, in which the filter corrects the state by combining the new measurement and the filter prediction weighted by their certainties. This filter, and its variants, are a common class of filter-based methods \cite{allotta2016new}, and are widely used due to their ability to produce better state estimates than those based on a single measurement. However, the KF is limited to linear functions for the state transition as well as the observation model. In our case, this reduces our ability to correctly track objects that can have non-linear motions, such as accelerations. The Extended Kalman Filter (EKF) overcomes this constraint by linearizing these functions, but it is often difficult to tune a single filter for all the motion modalities we encounter for each object. In this paper, we use the Interacting Multiple Model (IMM) \cite{Genovese2001TheIM} algorithm because it overcomes these limitations by tracking with multiple models concurrently and fusing their predictions weighted by their confidences. Furthermore, the IMM has been shown to offer performance similar to the best motion model.

\subsection{Tracking-by-Detection}
Most recent work on Multi-Object Tracking (MOT) utilize the tracking-by-detection paradigm \cite{zhang2008global,li2009learning,kuo2010multi,kim2012online,xiang2015learning,lenz2015followme,milan2017online,sadeghian2017tracking,schulter2017deep,farazi2017online,rangesh2019no} where detections are provided each time-step by a detector, and tracking is performed by linking detections across time.
As a result, the task of object tracking becomes a data association problem.
Most tracking-by-detection methods solve the association problem in one of two ways, either in an online (step-wise) fashion \cite{li2009learning,kuo2010multi,kim2012online,xiang2015learning,milan2017online,sadeghian2017tracking,farazi2017online,rangesh2019no} or in an offline (batch-wise) manner \cite{zhang2008global,lenz2015followme,schulter2017deep}.
Online methods associate new detections at each time-step to the existing tracks, and the association is posed as a bipartite graph matching problem.
On the other hand, offline methods often consider the entire sequence, and data association is cast as a network flow problem.
Online methods are appropriate for real-time applications like autonomous driving where offline approaches are well-suited for offline tasks like video surveillance.
In this work, we leverage a step-wise approach as we are interested in real-time autonomous navigation where computation efficiency is as important as accuracy.

\subsection{Classical Association Techniques}
\label{baselines}
To solve the data association problem, incoming detections at the current timestamp need to be paired to existing objects from the last timestamp. To avoid matching in the entire measurement space, every detection that lies within a certain region, or \emph {gating region}, of an object is considered a candidate. A problem arises when multiple candidates fall within this region. A common way to solve this involves ranking each object-detection pair and then performing a bijective mapping. The bijective mapping forces each object to associate with only one detection. This mapping can be performed using common matching algorithms such as greedy best-first matching or the Munkre's algorithm \cite{munkres1957algorithms}.    

A common method for ranking each object-detection is to score each detection based on the proximity to the predicted object \cite{bar2009probabilistic}. Based on this, we consider three functions:
\begin{enumerate}
\item \emph{Intersection-over-Union (IoU) score:}
Many trackers use ranking functions based a measurement of overlap between predictions and detections \cite{bochinski2017high}. This score is defined as the ratio between the area of intersection and the area of union of the detection and predicted polygons.
\item \emph{${L_2}$ Distance:}
As part of the \emph{update} step, the IMM needs to compute the \emph{residual} or \emph{innovation}, which is the difference between the predicted detection and the new detection. The $L_2$ norm of the \emph{residual} can be used as a matching score.
\item \emph{Mahalanobis Distance:}
The IMM computes the \emph{gain} or \emph{blending} factor that determines the relative weight of the new detection in the \emph{update} step. This \emph{gain} is used to scale the \emph{residual} vector, and the $L_2$ norm of the resulting vector can be used as a matching score.
\end{enumerate}

\subsection{Learned Association Techniques}
More recently Recurrent Neural Networks (RNNs) have been used for association \cite{farazi2017online,milan2017online,sadeghian2017tracking}, which motivates our use of a RNN for association in this work.
However, our proposed method and the previous work utilize RNNs in different ways.
\cite{farazi2017online} uses a single Long-Short Term Memory (LSTM) to associate all detections to all tracks.
However, it requires the number of objects to be fixed and known beforehand, which is not feasible for autonomous driving in urban environments.
\cite{milan2017online} uses an LSTM to estimate the affinity matrix between all detections and tracks one row at a time.
Most similar to our approach is the work of Sadeghian et al. \cite{sadeghian2017tracking}, who use three separate LSTMs to model the appearance, motion, and interaction of the tracked objects over time.
Each track has its own memory for each of the LSTMs, and appearance, motion, and interaction features are extracted for each detection using a set of Convolution Neural Networks (CNNs).
The output of the LSTMs and the CNNs are fed into a multi-layer neural network to estimate the likelihood that the detection should be associated to the track.
Unlike \cite{sadeghian2017tracking}, our proposed method uses a single LSTM to model multimodal features of an object over time.
Furthermore, in addition to an association probability, our approach predicts a score for each possible match in order to improve association in heavily crowded scenes, and we estimate the state of the object to improve tracking.
Finally, our method tracks objects in 3D where \cite{sadeghian2017tracking} tracks objects in the 2D image plane.

\subsection{3D Object Tracking}
The vast majority of the previous work performs object tracking in the image plane \cite{zhang2008global,li2009learning,kuo2010multi,kim2012online,xiang2015learning,lenz2015followme,milan2017online,sadeghian2017tracking,schulter2017deep}.
However, to autonomously navigate a vehicle through the world, we need to reason about the environment in 3D or from a bird's eye view.
Furthermore, the bird's eye view is a natural representation for fusing multiple sensor modalities like LiDAR, camera and RADAR.
Rangesh et al. \cite{rangesh2019no} extends \cite{xiang2015learning} to the bird's eye view to track vehicles.
In \cite{rangesh2019no}, vehicles are detected with an image-based detector and localized in the bird's eye view using a flat ground assumption or with 3D measurements from LiDAR when available.
The life-cycle of tracks is handled through a Markov Decision Process (MDP) where the policy is learned, and tracks are associated with detections using a Support Vector Machine (SVM).
In this work, our proposed method is capable of fusing detections from various sensing modalities including LiDAR and image-based detectors.
Furthermore, we associate objects across sensors and time using a RNN.

\begin{figure*}[ht]
  \includegraphics[width=1.0\textwidth]{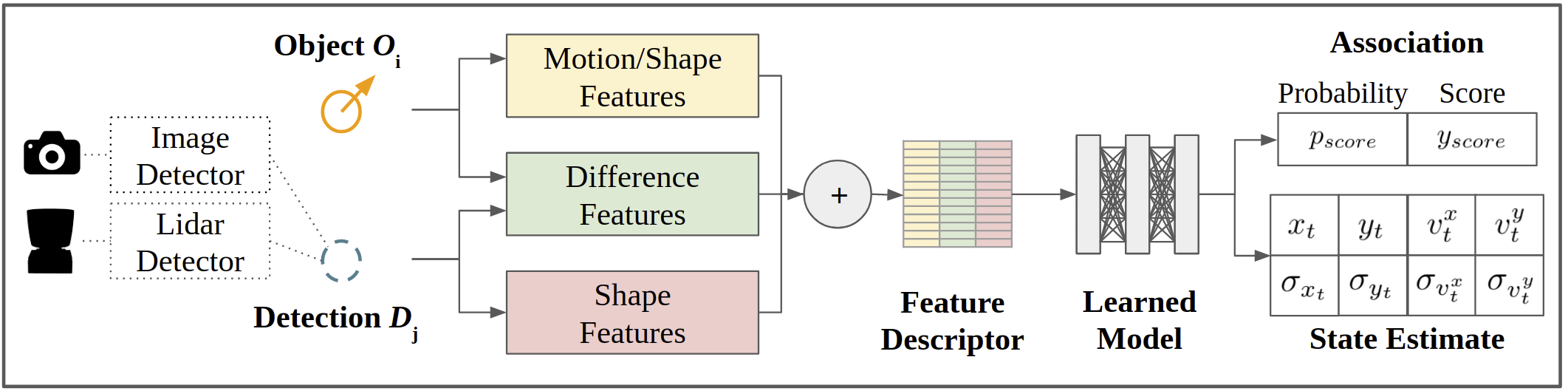}
  \caption{Proposed architecture of the learned association and state estimation model. We perform feature extraction for each candidate pair and learn whether the pair is a true association and the posterior state estimate of the object with uncertainties. We learn a probability of association and an association score to break ties between multiple competing candidates. We show that learning both a probability and a score are beneficial to the task of association, as well as the learned posterior state estimate improves overall tracking performance.}
  \label{fig:association_overview}
\end{figure*}

\section{Proposed Method} \label{sec:proposed_method}

The overall architecture of the system is depicted in Fig. \ref{fig:system_overview}. We use detections generated at each time step independently from LiDAR and camera sensors. To generate detections from LiDAR, we use LaserNet \cite{lasernet}, and the detections from the camera sensors are generated using RetinaNet \cite{lin2017focal}. Similar to \cite{rangesh2019no} and \cite{song2015object}, the image-based detections are then augmented with a range estimate by projecting the LiDAR points in the image plane and using the median range value of the points associated to create 3D bounding boxes. 

The proposed method is depicted in Fig. \ref{fig:association_overview}. During inference, the model takes an object-detection pair as input, and produces its association and state. For each object, we generate a set of potential association candidates with a corresponding score. The set of potential association candidates is created by predicting an association/mis-association probability for every pair. If the probability of association is higher than mis-association, then we add the pair to our set of potential association candidates. Afterwards, we perform greedy assignment based on the predicted score to create unique object-detection associations. We refine the detections with our predicted state estimate before using them as observations in the IMM. 

After updating the state for objects, we need to prune our existing hypothesis set of objects that are currently alive in the scene. Objects that have not been observed for more than $\tau$ time-steps are removed from the scene. For objects that have not been observed for $\leq \tau$ time-steps, we do a ballistic rollout of the objects to the next timestamp.

In the following sections, we describe in detail feature extraction from detection-object pairs, the network architecture, the multi-task loss function and the ground-truth association used during training.

\subsection{Feature Extraction}

We extract three different types of features: shape, motion and difference features. The shape features include polygon length, width, height and center coordinates. The motion features include the object's previous and predicted state. The difference features include difference in predicted object position and the detection position. We also use the timestamp and detector confidence as input to the model. While we could use a separate network for feature extraction or use the features from the internal activation layers of the detectors, we decided to utilize these lightweight features in order to keep our method real-time and sensor-agnostic.

\begin{figure}[t]
\centering
  \includegraphics[width=0.5\textwidth]{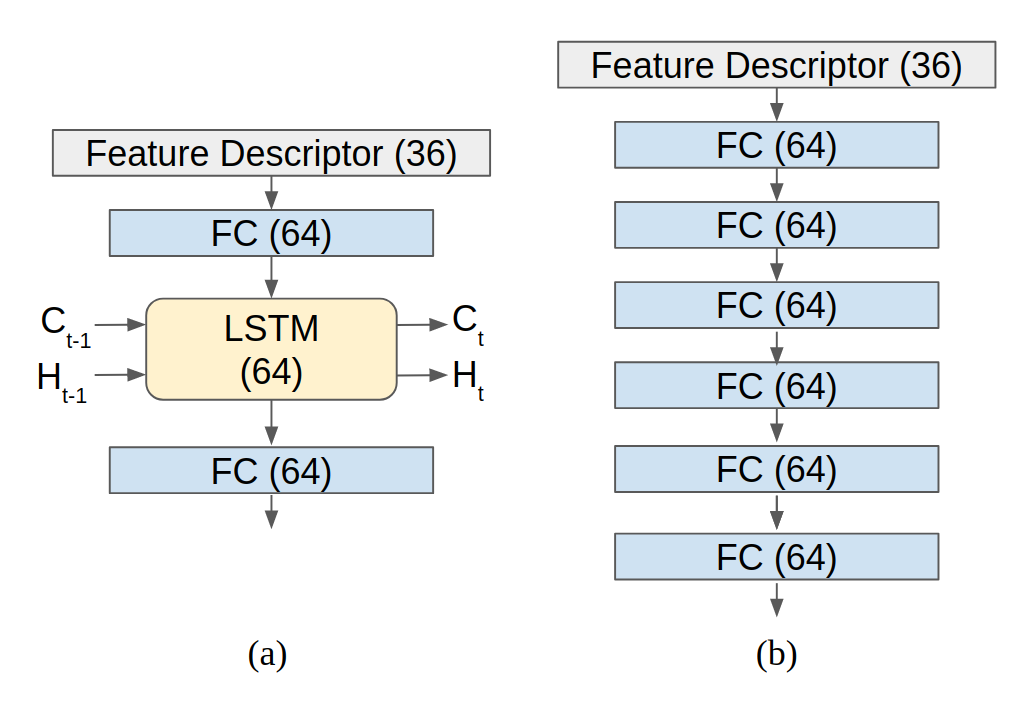}
  \caption{Network architecture used for LSTM and MLP networks. (a) For the LSTM, we use a single LSTM cell with 64 hidden units and a single layer fully-connected encoder-decoder. The network takes the feature descriptor, the cell state ($C_{t-1}$) and the hidden state ($H_{t-1}$) for the object as input to produce association outputs and new cell  ($C_{t}$) and hidden states ($H_{t}$). (b) For the MLP, we use six fully connected layers with 64 units each.}
  \label{fig:lstm_mlp_networks}
\end{figure}

\subsection{Learning Joint Association and Tracking}
The learned model produces association probabilities, scores and state estimates. For this work, we implement a single-cell LSTM as well as a Multi-Layer Perceptron (MLP). Both network architectures can be seen in Fig. \ref{fig:lstm_mlp_networks} and we compare the performance of each in Section \ref{recurrent_mlp_comparison}.

To learn association and tracking jointly, we utilize a multi-task loss. For the task of association, we propose learning a unique training target comprised of an association probability and score. The association probability is framed as a binary classification problem in which we try to categorize candidates as associations or mis-association. The association probability is used to identify a list of potential candidates that could potentially be associated. Furthermore, the score is used to rank associations, when there are more than one potential candidates for association. The loss function for the association task is defined as,
\begin{equation}
    \ell_{assoc} = \ell_{prob} + w_{score} \cdot \ell_{score}\text{,}
\label{eqn:association_overall}
\end{equation}
where $\ell_{prob}$ is the binary cross entropy used to learn the association probability, $\ell_{score}$ a $L_2$ loss on the regressed score, and $w_{score}$ is used to weight the two losses.

In addition to learning association, we learn a posterior state update for the object. 
The state of the object at time $t$ is defined as follows:
\begin{equation}
    \boldsymbol{s}_t = [x_t, y_t, v_{t}^x, v_{t}^y]     
\end{equation}
\begin{equation}
    \boldsymbol{\sigma}_t = [\sigma_{x_t}, \sigma_{y_t}, \sigma_{v_{t}^x}, \sigma_{v_{t}^y}]
\end{equation}
where $(x_t, y_t)$ is the position of the object, $(v_{t}^x, v_{t}^y)$ is the velocity of the object, and $(\sigma_{x_t}, \sigma_{y_t}, \sigma_{v_{t}^x}, \sigma_{v_{t}^y})$ are the corresponding standard deviations.
We learn the state using the following loss \cite{kendall2017uncertainties}:
\begin{equation}
\ell_{state} = \sum_{i} \left(\frac{\left({s}_{t,i} - {s}^*_{t,i}\right)^2}{2{\sigma}^2_{t,i}} + \log {\sigma}_{t,i} \right)
\label{eqn:state-loss}
\end{equation}
where ${s}_{t,i}$ is the $i$-th element of the state vector at time $t$, ${\sigma}_{t,i}$ is the corresponding standard deviation, and ${s}^*_{t,i}$ is the ground-truth state. The total multi-task loss is
\begin{equation}
    \ell_{total} = \ell_{assoc} + w_{state} \cdot \ell_{state}\text{,}
\end{equation}
where  $w_{state}$ is used to weight the relative importance of the two tasks.

\subsection{Training Procedure} \label{sec:training}
For training the network for association and tracking, we use a dataset with time-consistent IDs for labels. To provide direct supervision for the association task, we require a function that maps a candidate object-detection pair to a binary value indicating a true or false association, along with a score.

Given a set of detections ${\mathcal{D}_t = \{D_t^1, D_t^2, \ldots, D_t^N\}}$ at time~$t$ and a set of objects ${\mathcal{O}_{t-1} = \{O_{t-1}^1, O_{t-1}^2, \ldots, O_{t-1}^M\}}$ from time~$t-1$, the goal of ground-truth association is to define a mapping $f : \mathcal{O}_{t-1} \mapsto \mathcal{D}_t$ using the labeled data $\mathcal{L}_{t-1}$ and $\mathcal{L}_{t}$ at time $t-1$ and $t$. To handle the case where the object does not match to any detection, a null detection is added to $\mathcal{D}_{t}$. For each object $O_{t-1}^i \in \mathcal{O}_{t-1}$, we first identify the label $L^j_{t-1} \in \mathcal{L}_{t-1}$ with the maximum IoU overlap with the object. Afterwards, we find \textit{all} detections in $\mathcal{D}_t$  with an IoU~$\geq 0.1$ with the label $L^j_t$ at time $t$. All candidate detections are added to the training set as a true association, and their score is defined as
\begin{equation}
    y_{score} = \|\phi(L_{t-1}^j) - \phi(O_{t-1}^i)\|_{2} + \|\phi(L_{t}^j) - \phi(D_{t}^k)\|_{2}
\end{equation}
where $D_{t}^k$ is a candidate detection and $\phi(\cdot)$ computes the object's centroid. 

During inference the model will encounter mis-associations as well. Therefore, the model needs to learn to identify false associations. To accomplish this, we augment the dataset with examples of mis-associations. For every true association, $D_{t}^{k}$ and $O_{t-1}^{i}$, we identify all $D_{t}^{n} \in \mathcal{D}_t$ where $\|\phi(O_{t-1}^{i}) - \phi(D_{t}^{n})\|_2 < r$ and do not have an IoU~$\geq 0.1$ with $L_{t}^j$. We add a random subset of such examples to our dataset as false associations.

By predicting an association probability and a score, our method is robust to false positives due to duplicate detections. The probability allows us to identify all potential association candidates, including the true detection as well as false positives. The score then allows us to select the best candidate and discard the duplicate detections. In our experiments, we demonstrate the importance of predicting both.

Another advantage of breaking the problem of association into learning a probability and a score is that it eliminates the need for any engineered threshold to identify matches. Finding such thresholds can be challenging in the context of using different sources for detections with different error characteristics, e.g. image-based detections may have a higher range of uncertainty as compared to LiDAR detections. Besides, different VRU classes have different motion characteristics, e.g. bikes can move faster than pedestrians; therefore, different classes could have different scores. Our proposed method, considers all candidates with an association probability greater than the mis-association probability, and it identifies the best match with the score. As a result, we eliminate the need for any engineered thresholds.

\section{Experimental Results}

\subsection{Experimental Setup} 
We evaluate our method on the ATG4D dataset which contains 5,000 sequences in the training set, 900 sequences for the test set and 500 sequences for validation set. Each sequence is captured at 10Hz intervals. The data is collected using a Velodyne 64E LiDAR  along with a camera sensor. 
For the experiments in the paper, we generate detections as described in Section \ref{sec:proposed_method}. To reduce the detection-object pairs that we run inference for, we prune the list of all possible pairings based on a gating radius, $r$. This is common practice within tracking \cite{bar2009probabilistic} and makes the problem tractable by not considering impossible associations.  

For our experiments, we set $r=4$ m since it accommodates both slow moving pedestrians and fast moving bikes and $\tau=5$ for our object track life management. We set $w_{score}=0.02$ and $w_{state}=0.06$ while training models. Finally, the individual motion models in the IMM are designed to be adapted to the different motion modalities we encounter: \emph{static}, \emph{constant velocity}, and \emph{accelerating}.

\subsection{Evaluation Metrics}

We evaluate the performance of methods using standard multi-object tracking metrics \cite{bernardin2008evaluating, milan2016mot16} to compare tracking methods. These include evaluating the Multi-Object Tracking Accuracy (MOTA), Multi-Object Tracking Precision (MOTP), Mostly Tracked (MT), Mostly Lost (ML) and ID Switches (IDSW). However, these metrics fail to capture the quality of velocity estimates. Measuring the accuracy of the estimated velocity is imperative to evaluating tracking performance for trackers that are used by dependent systems to predict behavior. To resolve this gap in the metrics, we propose two new metrics: Multi-Object Tracking Velocity Error (MOTVE) and Multi-Object Tracking Velocity Outliers (MOTVE).

We define MOTVE as the average velocity error for all true positive objects. This is computed as
\begin{equation}
\text{MOTVE} =  \frac{ \sum\limits_{t=0}^T \sum\limits_{i=1}^M||v^i_t - \hat{v}^i_t||_2} {\sum\limits_{t=0}^T g_{t}}
\end{equation}
where $\hat{v}^i_t$ and $v^i_t$ refer to the estimated velocity of $i$-th object and its corresponding ground-truth label at time $t$ respectively. The number of object-label pairs present at time $t$ are denoted by $g_{t}$. 

We define MOTVO as the fraction of the object-label pairs where the velocity error is greater than a threshold,
\begin{equation}
\text{MOTVO} =  \frac{ \sum\limits_{t=0}^T \sum\limits_{i=1}^n \mathbf{1}[\|v^i_t - \hat{v}^i_t\|_2 > \nu]} {\sum\limits_{t=0}^T g_{t}}
\end{equation}
where $\mathbf{1}[\cdot]$ is an indicator function. For this evaluation, we set $\nu$ to $1$ m/s for pedestrians and $1.5$ m/s for bicyclists. This measures the number of egregious velocity errors and gives an indication about how robust the system is to producing velocity outliers.

\begin{table*}[t]
\centering
\caption{Comparison of Tracking Methods Across Multiple Sensor Modalities}
\resizebox{\textwidth}{!}{%
\begin{tabular}{c|c|ccccccccccccc}
\hline
\multirow{2}{*}{\begin{tabular}[c]{@{}c@{}}Sensing \\ Modalities\end{tabular}} & \multirow{2}{*}{Method}   & 
\multirow{2}{*}{MOTA $\uparrow$} &  
\multicolumn{2}{c}{MOTVO $\downarrow$} & 
\multicolumn{2}{c}{MOTVE $\downarrow$} & 
\multirow{2}{*}{FP $\downarrow$}  & 
\multirow{2}{*}{FN $\downarrow$} & 
\multirow{2}{*}{IDSW $\downarrow$} & 
\multirow{2}{*}{MOTP $\downarrow$} & 
\multirow{2}{*}{MT $\uparrow$} & 
\multirow{2}{*}{ML $\downarrow$} & 
\multirow{2}{*}{Frag $\downarrow$} \\
                  &        & & Ped         & Bike        & Ped         & Bike &         & &       &    &    &      \\ \hline
\multirow{4}{*}{LiDAR} 
                  &  \multicolumn{1}{l|}{IoU-based Association}          
                  &  $67.6486$ &  $3.572$  &  $2.377$  &  $0.170$  &  $0.287$    &  $394840$  &  $510918$  &  $44855$  &  $0.3527$  &  $0.389$  &  $0.164$  &  $39385$     \\
                  &  \multicolumn{1}{l|}{L2 Association}    
                  &  $67.9379$ &  $3.204$  &  $2.373$  &  $0.158$  &  $0.281$    &  $391298$  &  \textbf{508561}  &  $42092$  &  $0.3519$  &  $0.390$  &  $0.163$  &  $38850$\\
                  &  \multicolumn{1}{l|}{Mahalanobis Association}          
                  &  $68.4670$ & $2.956$  &  $2.041$  &  $0.157$  &  $0.271$    &  $370060$  &  $516321$  &  $37788$  &  \textbf{0.3466}  &  $0.386$  &  $0.165$  &  $40026$ \\
                  &  \multicolumn{1}{l|}{SDVTracker (Ours)}     
                  &  \textbf{68.9816} &  \textbf{2.199}  &  \textbf{1.633}  &  \textbf{0.131}  &  \textbf{0.248}    &  \textbf{362560}  &  $510970$  &  \textbf{35433}  &  $0.3475$  &  \textbf{0.391}  &  \textbf{0.162}  &  \textbf{38438}  \\\hline \hline
\multirow{4}{*}{\begin{tabular}[c]{@{}c@{}}LiDAR \\ + \\ Camera\end{tabular}}  
                  &  \multicolumn{1}{l|}{IoU-based Association}         &  $66.5809$ &  $4.236$  &  $2.549$  &  $0.192$  &  $0.295$    &  $416723$  &  $503642$  &  $51731$  &  $0.3586$  &  $0.384$  &  $0.167$  &  $43417$     \\
                  &  \multicolumn{1}{l|}{L2 Association}     &  $68.1027$ & $3.303$  &  $2.334$  &  $0.162$  &  $0.294$   &  $386467$  &  \textbf{497147}  &  $43991$  &  $0.3554$  &  \textbf{0.388}  &  $0.163$  &  $40572$  \\
                  &  \multicolumn{1}{l|}{Mahalanobis Association}   &  $68.6031$ & $3.056$  &  $2.118$  &  $0.160$  &  $0.287$    &  $366913$  &  $504202$  &  $39521$  &  $0.3498$  &  $0.385$  &  $0.165$  &  $41251$     \\
                  &  \multicolumn{1}{l|}{SDVTracker (Ours)}                     &  \textbf{69.4405}  &  \textbf{2.204}  &  \textbf{1.827}  &  \textbf{0.133}  &  \textbf{0.268}  &  \textbf{346651}  &  $504744$  &  \textbf{33118}  &  \textbf{0.3485}  &  \textbf{0.388}  &  \textbf{0.162}  &  \textbf{40008}  \\\hline
\end{tabular}%
}
\label{tab:baselines-table}
\end{table*}

\subsection{Performance Comparison}
 We compare our learned method for joint association and tracking to the classical association methods described in Section \ref{baselines}, due to their widespread use in filter-based tracking for real-time systems. We evaluate all methods on unimodal (LiDAR Only) and multimodal (LiDAR + Camera) configurations. All methods use the same IMM tracker. The results are detailed in Table \ref{tab:baselines-table}.
Our proposed SDVTracker significantly improves system performance over other methods for both sensor modalities. For the LiDAR only system, we see improvements such as a $16\%$ reduction in MOTVE, a $6.23\%$ reduction in ID switches and a $2\%$ reduction in false positives, over the next best method. Mahalanobis association has the best MOTP by 0.09 cm, but does not translate to better velocity estimates. This further demonstrates the need of metrics that measure higher order states when evaluating object tracking in 3D.

Furthermore, as more sensors are added to the system, we see an improvement in the overall MOTA and false negatives of methods. However, this comes at the cost tracking more objects, increasing the absolute number of velocity outliers. We show that our learned methods can better incorporate new sensor observations by reducing velocity outliers by $17\%$, ID switches by $16\%$ and false positives by $5\%$. While Mahalanobis association sees a degradation in performance by around $3.3\%$, our learned method sees an increase in velocity outliers by $0.2\%$, all the while tracking more objects.

\subsection{Impact of Recurrent Networks} \label{recurrent_mlp_comparison}
We implement two learned network architectures for our learned association and tracker. For the recurrent network, we train on truncated sequences of length 20. We compare the performance of a Recurrent Neural Network (RNN) to a feedforward Multi-Layer Perceptron (MLP) in Table~\ref{tab:tracking_ablation}. While both networks outperform classical association methods, we see a small increase in performance with the recurrent network.

\begin{table}[t]
\centering
\caption{Effect of Learning Joint Tracking and Association}
\resizebox{0.48\textwidth}{!}{%
\begin{tabular}{c|c|c|cccc}
\hline
Network & IMM & Learning State & MOTA $\uparrow$ & MOTVO $\downarrow$ & MOTVE $\downarrow$ & IDSW $\downarrow$ \\ 
\hline
  MLP   &  $\checkmark$ &       & $69.2221$     & $2.448$     & $0.1446$      & $37594$     \\
  MLP   &  $\checkmark$ & $\checkmark$     & $69.3863$     &     $2.385$  & $0.1413$      & $34698$  \\ \hline
  LSTM  &  $\checkmark$ &        & $69.2877$  & $2.428$      & $0.1419$      &    $35862$  \\   
  LSTM  &   & $\checkmark$      & $69.3971$     & \textbf{2.240}      &  $0.1528$     &  $34031$  \\
  LSTM  &  $\checkmark$ & $\checkmark$      & \textbf{69.4405}     & $2.292$      &  \textbf{0.1393}     &  \textbf{33118} \\
\hline
\end{tabular}
}
\label{tab:tracking_ablation}
\end{table}

\begin{table}[t]
\centering
\caption{Effect of Learning Probability and Score}
\resizebox{0.48\textwidth}{!}{%
\begin{tabular}{c|cccc}
\hline
Association Output        & MOTA $\uparrow$ & MOTVO $\downarrow$ & MOTVE $\downarrow$ & IDSW $\downarrow$ \\ \hline
Probability Only      & $69.1837$     & $2.544$      & $0.1466$      & $35419$      \\
Score Only            &  $69.3618$    & $2.551$   & $0.1448$     &    $39461$  \\
Probability and Score  & \textbf{69.4405}     & \textbf{2.292}      &  \textbf{0.1393}     &  \textbf{33118} \\ 
\hline      
\end{tabular}
}
\label{tab:target_ablation}
\end{table}

\subsection{Ablation on Joint Association-State Estimation}
To understand the impact of jointly learning association and state estimation, we trained a recurrent and a feed-forward network with and without including state estimation learning as a model output. The results are outlined in Table \ref{tab:tracking_ablation}. We see that regressing the state information improves performance for both network architectures. Further, we investigate how the model's learned state compares with the filtered IMM state. We see that while the model's learned state produces fewer velocity outliers, its average velocity and MOTA are worse compared to using the IMM, which motivates our hybrid method.

\subsection{Ablation on Score Regression}
We evaluate the effectiveness of learning both an association probability and a score, as discussed in Section \ref{sec:proposed_method}, in Table \ref{tab:target_ablation}. For the probability only model, we break ties between candidate detections based on the higher probability. For the score only model, we considered all scores below $0.1$ as candidate associations. Based on the results, we see that neither breaking ties with the probability or thresholding based on the score perform better than explicitly learning a probability and a score.  

\begin{figure}[t]
  \includegraphics[width=0.485\textwidth]{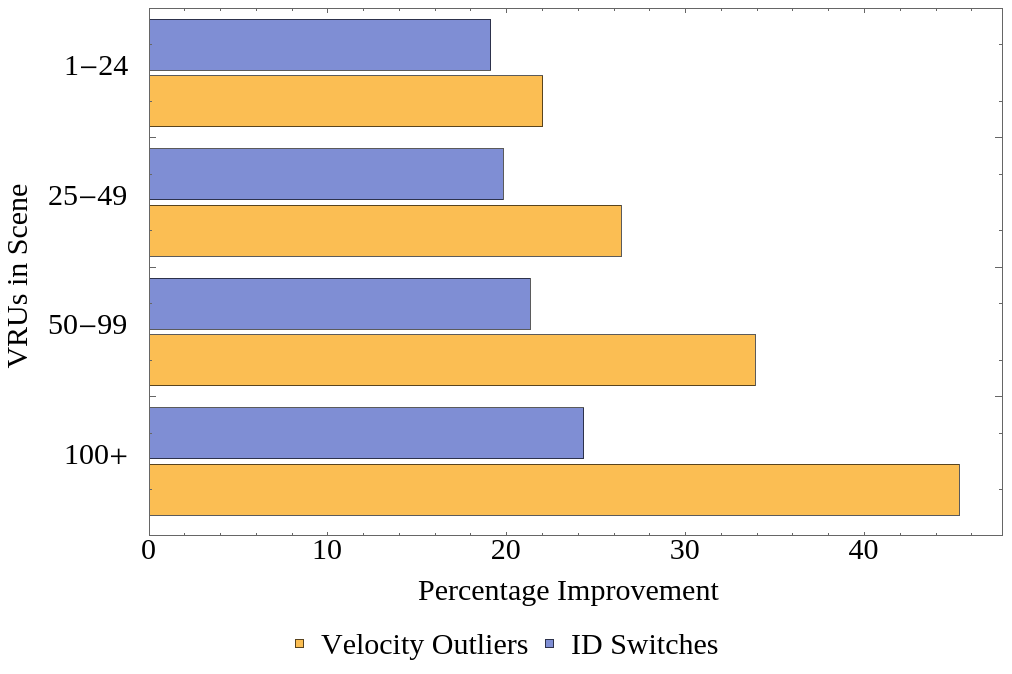}
  
  \caption{As the number of pedestrians in a scene grows, our method is increasingly more effective at reducing velocity outliers than engineered methods. Analysis was performed on over 900 scenes bucketed by the number of pedestrians across a 25s interval, with each bucket including at least 20 scenes. }
  \label{fig:crowd_stats}
\end{figure}

\begin{figure*}[t]
  \includegraphics[width=1.0\textwidth]{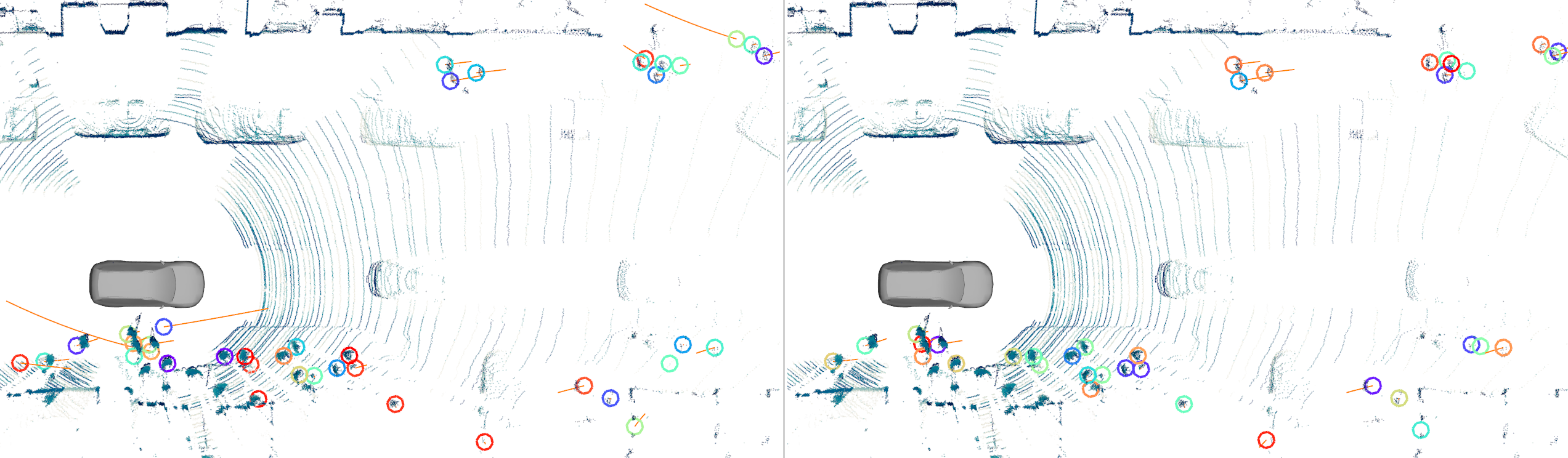}
  \caption{(left) Classical Mahalanobis association and tracking. (right) SDVTracker, our system for learned association and tracking, which shows fewer velocity outliers. Circles represent tracked VRUs and orange vectors represent velocity estimates. See attached supplemental material for video versions.}
  \label{fig:qualitative_comparison}
\end{figure*}

\subsection{Impact of Pedestrian Density}
In dense crowds, a mis-association can cause a tracked object to have poor velocity estimates, which degrades system performance. Fig. \ref{fig:crowd_stats} examines the performance of SDVTracker as the number of pedestrians in a scene is increased in terms of ID switches and velocity outliers. As pedestrian density increases, our proposed method performs better than hand-engineered association on both metrics. In scenes with 100+ pedestrians, the learned model reduces poor velocity estimates by 45\%, demonstrating our learned model approach scales better than classical methods. 

\subsection{Runtime Performance}
\begin{figure}[tbh!]
  \centering
  \includegraphics[scale=0.25]{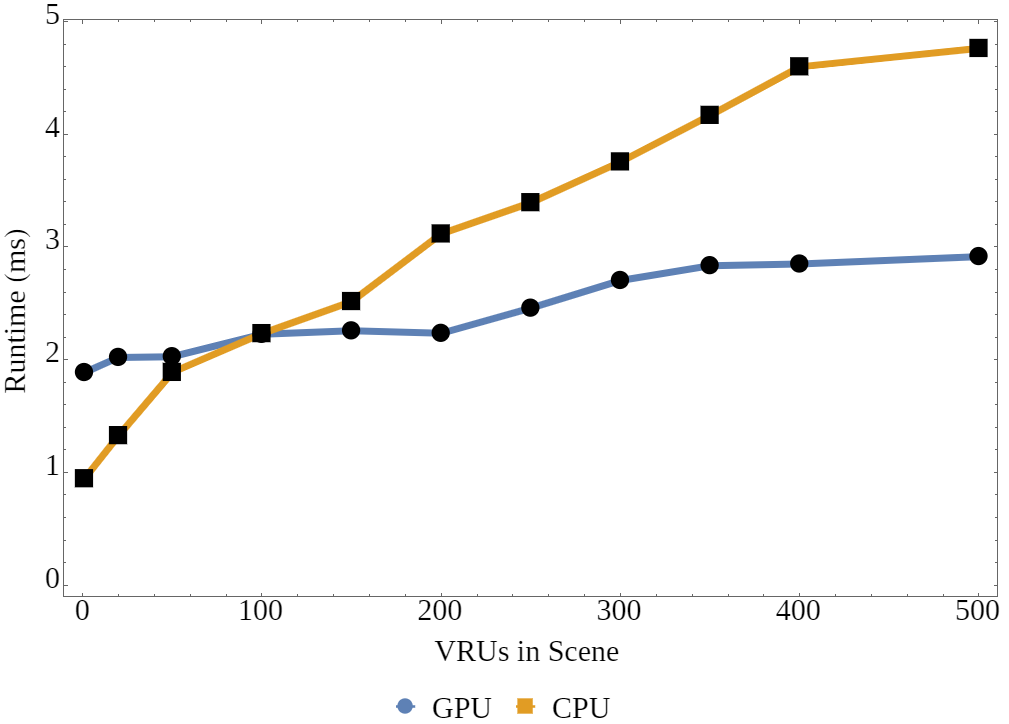}
  
  \caption{Model inference runtime on CPU and GPU as a function of the number of actors in a scene. The model scales approximately linearly with the number of actors and for a typical scene with 100 actors runs under 2.5 ms on CPU.}
  \label{fig:runtime}
\end{figure}

We show the runtime performance of the system in Fig. \ref{fig:runtime}, evaluated on a four core Intel i7 CPU and a NVIDIA RTX 2080Ti GPU. We see that model runs under 5 ms for 500 actors on a CPU and under 3 ms on a GPU. It is interesting to note that for scenes with less than 100 VRUs, it is faster to run on CPU than using a dedicated GPU. 

\subsection{Qualitative Performance}
Fig. \ref{fig:qualitative_comparison} shows representative output of the classical Mahalanobis association and tracking compared to SDVTracker on a typical scene with VRUs. We see fewer velocity outliers, which yields better self-driving vehicle performance. Please refer to the provided supplemental video to see the SDVTracker in operation. 

\section{Conclusion and Future Work}

We presented SDVTracker, a method for learning multi-class object-detection association and motion state estimation. We demonstrate that this algorithm improves tracking performance in a variety of metrics. In addition, we introduce new tracking metrics important in self-driving applications that measure the quality of the velocity estimates and show that SDVTracker significantly outperforms the compared methods. Furthermore, we demonstrate that SDVTracker generalizes to multiple sensor modalities, increasing recall with the addition of the camera sensing modality. Finally, we show this method is able to handle scenes of 100 actors under 2.5 ms, making it suitable for operation in real-time applications.

The performance of the learned state obtained directly from the LSTM was similar to the one obtained by the IMM, opening a door for new experiments to potentially remove the IMM from the algorithm while maintaining the performance. We plan to also augment the algorithm to learn the object life policy, controlling when to birth new objects and reap old ones. Finally, we further plan to extend SDVTracker by adding additional sensors, such as RADAR, to the system. 

\section{Acknowledgements}

We would like to acknowledge and thank several people at Uber ATG who have made this research possible. We thank Blake Barber, Carl Wellington, Chengjie Zhang, David Wheeler, Gehua Yang, Kyle Ingersoll, Narek Melik-Barkhudarov, and Ralph Leyva for their support.


\addtolength{\textheight}{-10cm}

\bibliographystyle{IEEEtran}
\bibliography{references}

\end{document}